\documentclass{article}

\usepackage{arxiv}

\usepackage[utf8]{inputenc} % allow utf-8 input
\usepackage[T1]{fontenc}    % use 8-bit T1 fonts
\usepackage{hyperref}       % hyperlinks
\usepackage{url}            % simple URL typesetting
\usepackage{booktabs}       % professional-quality tables
\usepackage{amsfonts}       % blackboard math symbols
\usepackage{nicefrac}       % compact symbols for 1/2, etc.
\usepackage{microtype}      % microtypography
\usepackage{lipsum}
\usepackage{graphicx}
\usepackage{amsmath}

\title{Group Activity Recognition via Dynamic Composition and Interaction}

\author{Youliang Zhang\\
Wuhan University\\
\texttt{ zyliang2303@163.com}
\and
Zhuo Zhou\\
Wuhan University\\
\texttt{ 305272@whut.edu.cn}
\and
Wenxuan Liu\\
Wuhan University of Technology\\
\texttt{ lwxfight@126.com}
\and
Danni Xu\\
The National University Of Singapore\\
\texttt{  dannixu.gracie@foxmail.com}
\and
Zheng Wang\\
School of Computer Science, 
Wuhan University\\
\texttt{  wangzwhu@whu.edu.cn}
}

\begin{document}
\maketitle
\begin{abstract}
   Previous group activity recognition approaches were limited to reasoning using human relations or finding important subgroups and tended to ignore indispensable group composition and human-object interactions. This absence makes a partial interpretation of the scene and increases the interference of irrelevant actions on the results. Therefore, we propose our DynamicFormer with Dynamic composition Module (DcM) and Dynamic interaction Module (DiM) to model relations and locations of persons and discriminate the contribution of participants, respectively. Our findings on group composition and human-object interaction inspire our core idea. Group composition tells us the location of people and their relations inside the group, while interaction reflects the relation between humans and objects outside the group.    
   % Based on this, we proposed a novel XXX to effectively perform group activity recognition with keypoint only modality, by modeling the human-object interaction and human relations. 
   We utilize spatial and temporal encoders in DcM to model our dynamic composition and build DiM to explore interaction with a novel GCN, which has a transformer inside to consider the temporal neighbors of human/object. 
   % In addition, to capture the human-object interactions, we build DiM with a novel GCN, which has a transformer inside to consider the temporal neighbors of human/object.
   Also, a Multi-level Dynamic Integration is employed to integrate features from different levels. %(such as interaction, relation and group features) 
    We conduct extensive experiments on two public datasets and show that our method achieves state-of-the-art. %danni: achieve or outperform?  the performance of our method is comparable to state-of-the-art methods, which use more inputs. 
\end{abstract}
%%%%%%%%% BODY TEXT
\section{Introduction}
%composition interaction
Group Activity Recognition (GAR) refers to recognizing activity scenes containing multiple people over time, aiming to determine what activities these people are engaged in~\cite{deng2016structure, wang2018appearance, tran2015learning, simonyan2014two, li2021groupformer, zhou2022composer}. Benefiting from the success of explorations about relation inferring and graph struct processing~\cite{azar2019convolutional,ibrahim2018hierarchical,bagautdinov2017social,li2017sbgar,shu2017cern,wu2019learning}, GAR has achieved excellent progress recently. However, as the focus is to distinguish the group category, inferring the connection between the individual and the whole remains a debatable challenge.

%danni: equally这个词是不是准确， sub-regions的影响是什么意思
% To address the issue,

%这里引出composition有些奇怪 现有方法概括粒度不对。0307
% Usually, the aforementioned methods consider all individual actions equally while ignoring the diverse influence of some more important sub-regions.  
%平均或者静态的对待frame上的每一个sub-group？没有考虑sub-group会随着之间改变relation和空间变化？？好难啊
%实际上是变化的结构导致每个人的importance也是变化的
%问题要怎么描述呢？rigid/fixed/inflexible importance？
%现有的方法缺点 一层套一层 然后点出问题 %每一个时刻的关注度？都是相同的？
%: 这里我理解的是，在时序上，key person和sub-group都是在变幻的，可能从一个人变成另一个人，也可能从左边变到右边，因此需要考虑动态的移动和改变
%.先把dynamic 概念拎出来？
Existing methods~\cite{ehsanpour2020joint,nakatani2021group,li2021groupformer} introduce the sub-group or key person to select the essential part representing the globally trusted representation, which is static in each moment. Nevertheless, a recent study~\cite{YanSYTT22} suggests that key participants should have continual movement throughout the process. In other words, it is necessary to concentrate on the \textbf{dynamic}
motion to understand group activities effectively. 

\begin{figure}
    \centering
    \label{fig:inter_motivation} \includegraphics[width=\columnwidth]{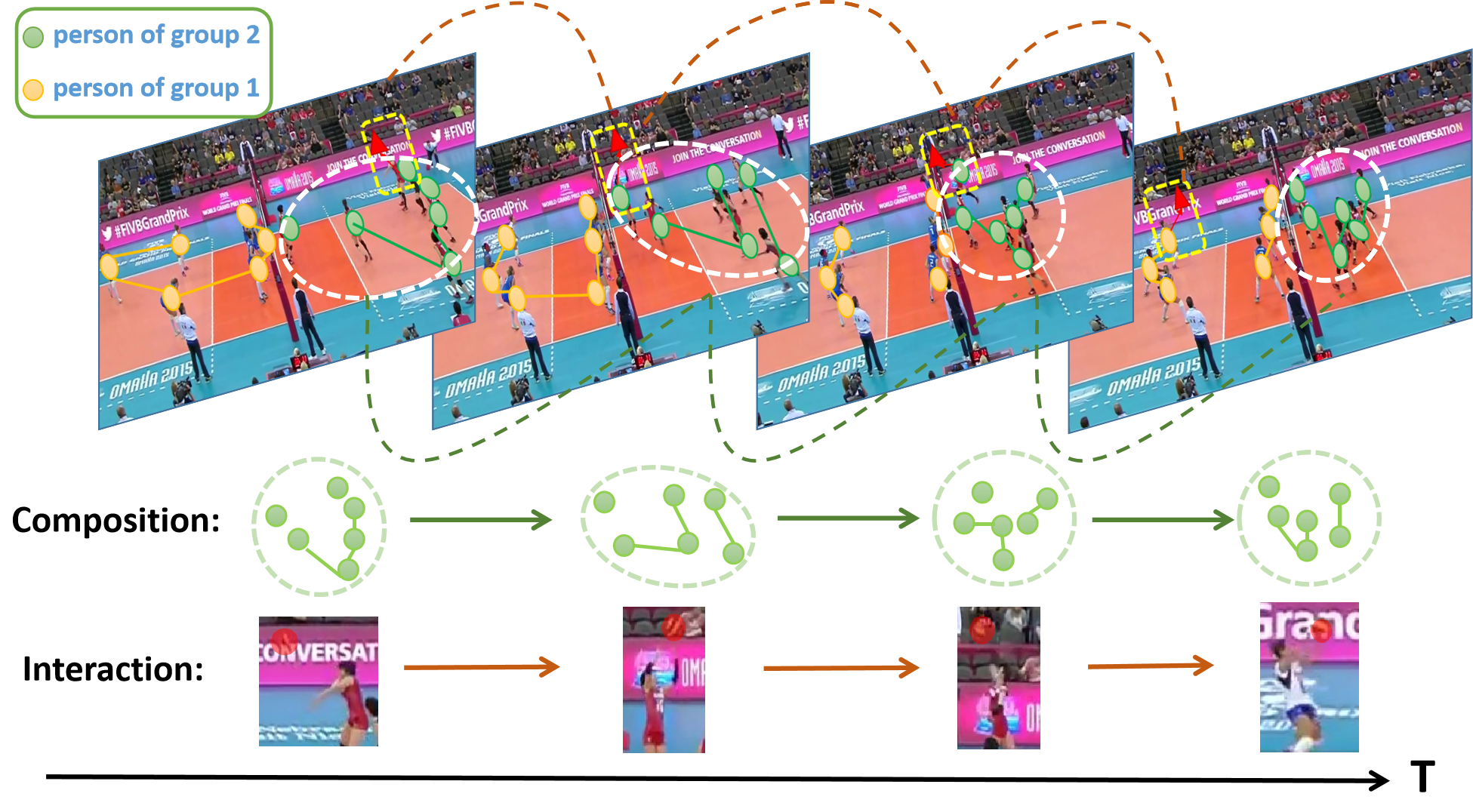}
    \caption{Motivation of our proposed method. The dots represent the location of people and connecting lines indicate human relations. Human-object interaction is marked using yellow boxes. It shows that human-object interaction and group composition change dynamically over time. }
    %\caption{(a) shows human-object interaction and group composition. The dots represent the location of people and their connecting lines indicate human relations. It shows that the interaction and composition change simultaneously. (b) is designed to show this association, where the X-axis (Green dot graph) represents a different kind of composition of the group, different curves represent different interactions, and the Y-axis represents the corresponding probabilities of interaction and composition. The yellow arrows indicate the most likely composition for an interaction.}
\end{figure}

%spatial distribution with relations
Indeed, we observe that human motion is frequently present and changed in various other forms. As illustrated in Figure~\ref{fig:inter_motivation}, we have highlighted the person's movements along the temporal dimension. Take group 2 (circled in Figure~\ref{fig:inter_motivation}) as an example, the spatial distribution changes from dispersion to aggregation and the relationship between humans express diversely at different moments. The two perform flexibly during the whole activity processing, defined as \textit{composition.} %Through the significance investigation, 
Based on the above observation, important clues have emerged that the dynamic composition provides more functional characterization to understand the activity than other static contributions.

This motivates us to construct the Dynamic-composition Module (DcM), which considers the relation information and highlights the dynamic physical distribution of groups. DcM utilizes spatial and temporal encoders in a novel way to capture the group composition and their changes from a global perspective, which is more conducive to our knowledge of the whole scene. 
%

%lack the ability to capture the dynamic changes of group composition, including human relations and group distribution, and the dynamic changes of actor importance. Unaware of those dynamic change results in the wrong focus on some immaterial actors and then influences the recognition accuracy, since different person have different contribution to the group. % 
% liang: 这里是不是需要提前介绍一下composition的概念？danni:这里最好再描述一下两个缺陷。（从文中好像是两个缺陷，但是现在的语句好像揉在一起了，也看不出其中是什么关系～）1. 说明动态的compostion现象是什么, 并加个定语话解释composition的意思 2. 加一句话说明该动态性会如何影响GAR的的判断，3. 一句话说明immateiral actors的现象+忽视该现象可能会导致GAP准确性有偏差。
% introduce sub-group or key actor to represent which sub-region is more important and which part of the scene requires our special attention. 

%To cope with the above questions, we did the statistical analysis and visualization about the dynamic relationship between group activity and group composition. The results (see Figure~\ref{fig:inter_motivation}) show that 
% Based on our visual observation and statistical results shown in Figure\ref{fig:inter_motivation}, we find that : 1) the group composition is dynamic during the competition, and 2) different person of a group have unequal contributions to the group activity recognition.

%Based on these findings we propose a novel DynamicFormer with a Dynamic composition Module (DcM) and a Dynamic interaction Module (DiM). DcM focuses on the global picture, emphasizing the overall features of the group, while DiM pays attention to the local area and provides us with information about the key persons. The details are as follows.

%As shown in Figure\ref{fig:inter_motivation} (a), we highlight each actor's location and the relation with others at different frames. We visualize group composition in Figure\ref{fig:inter_motivation} (a) with a combination of points and lines, representing the locations of person (Distribution) and the relationship between humans (Relation) respectively. From this figure we can find that the set of locations of people in a subgroup presents a variety of dynamics.%zhouz-- We visualize group composition in Figure\ref{fig:inter_motivation} (a) which shows the change in the position of people over time.

%这里是动态人物交互 为什么人物交互重要
 %danni: explain why human-object interaction can solve the problem~
As the composition displays a dynamic status, another phenomenon is concomitantly shown in Figure~\ref{fig:inter_motivation}. There are low-association outliers, such as ``standing'' persons, irrelevant to the current scene. Previous work \cite{DengVHM16} borrows the temporal context to reason the related scene and points out that the involved person. It lacks spatial analysis and may pose an unreliable prediction. As we can see from the figure, the categories of key participants from left to right are digging$\rightarrow$setting$\rightarrow$spiking$\rightarrow$digging. The key participants revolve around the ``volleyball" in this sports scenario, and other persons far away from the object may cause interference.

We need to weed out the distraction using the \textit{interaction} between person and volleyball. Human-object interaction (HOI) can construct the contextual details presented in the video, which motivates us to leverage the HOI information to solve this problem. To this end, we present another critical module, Dynamic-interaction Module (DiM), to capture the vicissitudes of human-object relations. It mainly uses a novel GCN, with a transformer inside, to process human and object features, exploring whether there is an interaction and how strong this interaction is. The DiM restrains the negative information that may be brought about by adding some station positions and including the positive effect of human interaction. 

% Interaction is an essential part of the scene, but it seems to be only a partial representation, missing the description of the overall composition. 

%In (b), our statistical analysis shows a tight connection between group composition and interaction, which implies % Danni: how does the connection imply the guiding? explain with the figure~ 
%that the combination of them can better guide global activity. This finding not only shows exploring group activity through interaction is much meaningful, meanwhile, it makes the collaboration between DcM and DiM feasible. Specifically, DiM points out the localities that DcM should focus more on, while DcM provides a global dynamic perception for DiM.
In this paper, we explore and utilize the association between composition and interaction to present a global framework, DynamicFormer, of the overall processing. First, we use DcM to process the human features extracted from keypoint information and generate our dynamic composition features. Also, a DiM is developed by GCN to model human-object interaction, considering the temporal neighbors of a human or object with a transformer. Then, a Dynamic Integration based on the transformer is used to integrate interaction, composition, and other useful information, since composition and interaction features complement each other and provide different perspectives on the scene, and we finally get a comprehensive and enlightening global characterization for GAR.

Our contribution is mainly threefold:
\begin{itemize}
    \item To explore the group composition changes across the spatio-temporal series and the unequal importance of different persons for group activity recognition. We propose the Dynamic-composition Module (DcM) and Dynamic-interaction Module (DiM). Precisely, the DcM models the dynamic group composition to get comprehensive features, and the DiM reduces the negative impact of the unimportant person while highlighting those vital persons.
    \item We analyze the connection between dynamic group composition and human-object interaction and propose DynamicFormer, which can complement to obtain comprehensive group features.
    % The proposed DynamicFormer explores the motion patterns in the scenario of diverse sub-group and presents the progressive relationship, %danni: involve many new terms: motion patterns, progressive relationship. what does progressive relationship mean?~
    % i.e., from the individual human-object interaction to sub-group composition to the relevant orders of group activity. % danni: Orders have not been introduced in the previous paragraphs. If it is important should add it to the previous part, otherwise delete it~
    \item To demonstrate the model’s strength, we perform extensive experiments on two widely used public datasets, Volleyball, and Collective datasets, and successfully achieve the performance of the state-of-the-art methods with only keypoint modality information. % danni: achieve or outperform? 
\end{itemize}

\section{Related Work}  
\subsection{CNN and GCN for Video Recognition}
CNN has achieved successful applications in image recognition, but it cannot be directly applied to video recognition, since it can not model the temporal information, which is vital for understanding videos.
To apply CNN to the video domain and extract temporal features, Ji~\cite{ji20123d} and Tran~\cite{tran2015learning} et al. propose and improve 3D CNN, which provides a good feature representation method and can be easily integrated with other models. 
% To better model useful information in videos
After that, Ibrahim et al.~\cite{ibrahim2016hierarchical} proposed an RNN-based model that uses CNN to extract features and RNN to utilize temporal information. 

\begin{figure*}[t]
    \centering
\includegraphics[width=\textwidth]{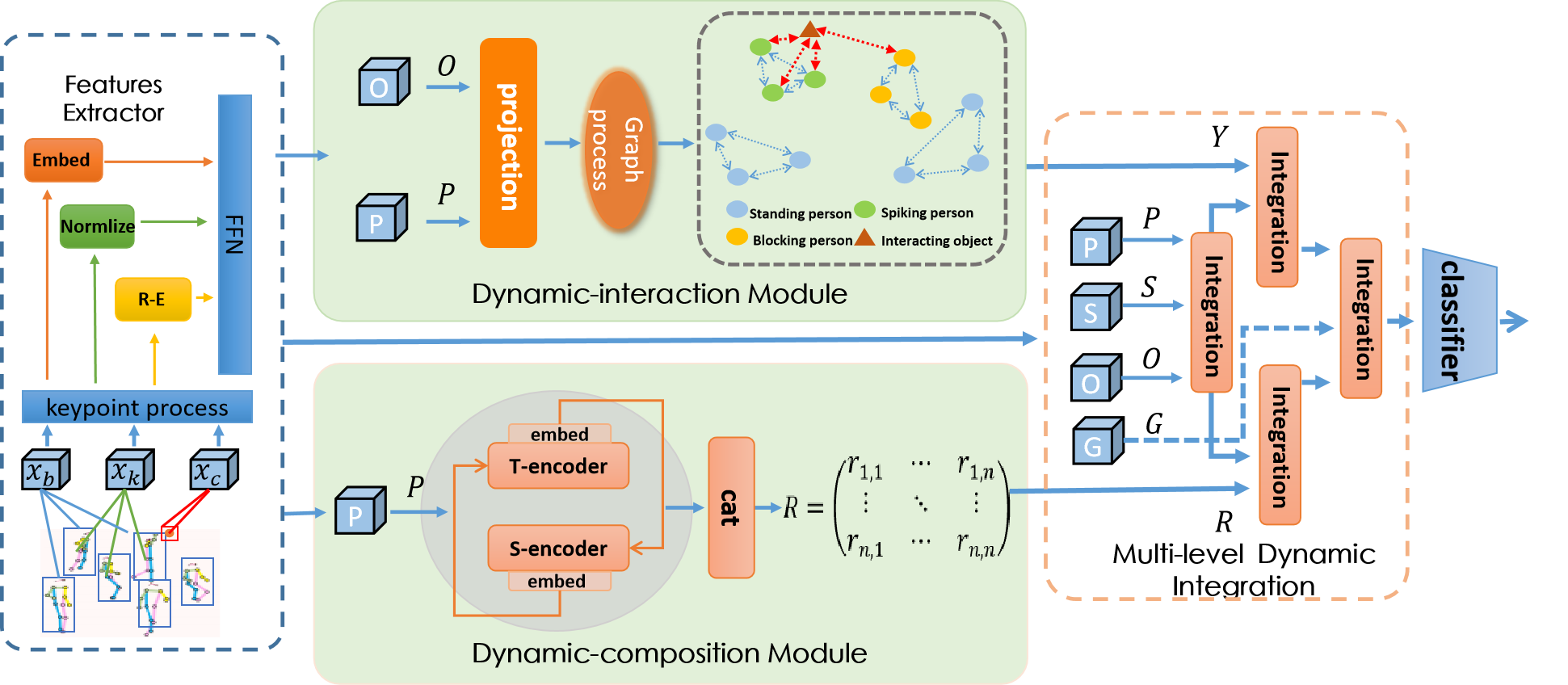}
    \caption{\label{fig:overview} Illustration of our proposed DynamicFormer. It contains four main components: 1) Our model takes human keypoints, bounding boxes and object coordinates as input, and the $k, o, p$ here represent keypoint, object, and person features respectively. 2) a DcM that takes human features as input to explore the relation and distribution information. 3) DiM uses an interaction GCN that takes object features and human features as input to explore the interaction information between humans and objects. 4) a multi-level integration transformer using the information of different levels and generating the final output for classifying.}
\end{figure*}

In recent years, Graph Convolutional Networks (GCN) attracts increasing attention, for it's ability to process structured data~\cite{kipf2016semi, kipf2018neural, wu2020comprehensive, xu2018powerful}. Wu et al.~\cite{wu2019learning} introduced GCN into group activity recognition, which combines CNN and GCN to explore relations between humans, by modeling multi-person scenes as a sparse graph with a temporal sampling strategy. ~\cite{hu2020progressive} combines graph with RNN and builds a semantic relation graph, representing each relation explicitly. However, these methods emphasize the representation of features and the construction of human relations in a group, lacking the exploration of dynamic group composition, which is really important since human relations and distribution change over time. Our method fully considers the dynamic composition with DcM, modeling not only dynamic human relations but also group distribution changes. 
% Many researchers have subsequently used GCN for modeling from different perspectives and have achieved many achievements. 
\subsection{Transformer}
After the emergence of Transformer~\cite{vaswani2017attention}, it quickly became a popular method in many fields of computer vision, for its ability to better capture long-term dependencies compared with RNN. Some studies explore the utilization of transformer in video action recognition, by collaborating it with graphs~\cite{wang2018videos} and LSTM~\cite{li2018videolstm}.
In the group activity recognition field, Gavrilyuk et al.~\cite{gavrilyuk2020actor} combine transformer with CNN for group activity recognition. ~\cite{li2021groupformer} utilizes transformer to build the bridge between the spatial and temporal information and obtained great performance. ~\cite{zhou2022composer} use a multi-scale transformer to model group activity recognition problem with keypoint modality-only information.

The mentioned methods utilize transformer to explore the temporal information when modeling human relations but ignore the meaningful human-object interaction that helps us to focus on the more important person. Through the collaboration with DcM, DiM explores interaction information and reveals which person is more important in global composition. Also, we use a multi-level transformer as an integration module to fuse composition, interaction, and other useful information.

% 这一段
% Previous methods ignore exploring the interaction between objects and humans, focusing only on human relations, which may lead to the absence of some important information. Moreover, human relations play an important role in determining the more important actions and people in the scene, but many methods just consider spatial information when inferring human relations. Unlike them, we adopt a Graph Convolution Network to explore the interaction information, which can be complementary to human relation information. 

\section{Methods}
We propose a novel model, DynamicFormer, to capture the dynamic motion pattern using relation and interaction information for group activity recognition involving multiple granularities, as illustrated in Figure \ref{fig:overview}. %We first describe how to obtain the feature representation from the keypoint and RGB modalities. Then, we elaborate on the Dynamic-compostion Module and Dynamic-interaction Module, respectively. And a multi-level Dynamic Ordered Integration is used to explore the levels of semantic and scene information to ultimately obtain reliable features for the classification of group activity.
  
    % an embedding and tokenizing block is used for collation and feature extraction of the input
\subsection{Feature Extractor}
Our DynamicFormer utilizes only keypoint modality information, which is not only lightweight but also effective. Given a video clip with a $T$-frames describing a specific group activity with $N$-persons, let $x_{b}, x_{k}, x{c}$ denote the human bounding box, human keypoint, and ball coordinate respectively. Our model takes $x_{b}, x_{k}$ and $x_{c}$ as input and obtains a number of representations with different semantic information below. The processing of this information includes simple grouping, normalization, difference, and tokenization. When processing the keypoint information, we calculate its coordinates relative to the center of the character, the absolute speed of the keypoint, the speed of the keypoint relative to the center of the character, the normalized speed and coordinates, etc., and process the object coordinates accordingly. Also, with the use of time and position embedding, we enhance the spatio-temporal representation ability of the extracted features.

\textbf{Object features} are extracted from ~\cite{perez2022skeleton}, where the features are provided in the form of object coordinates. We define object features as $\textbf{O}_t^e \in \mathbb{R}^D$ %$ \textbf{O}_{\textbf{t}}^{\textbf{e}} \in R^{D}$, 
where $\textbf{O}_t^e$ denotes the $e$-th Object at frame $t$, each of these contains information on relative position, absolute position, relative velocity, absolute velocity, and position embedding. $D$ is the dimension of that information, $E$ denotes the maximum number of objects in a video and $T$ denotes the total number of frames.

\textbf{Human keypoints} are extracted by HRNet from ~\cite{zhou2022composer} and we define it as $\textbf{S}_t^{n,k} \in \mathbb{R}^{D}$, where $\textbf{S}_t^{n,k}$ denotes the body joint of type $k$ in human $n$ at time $t$, $N$ and $K$ denote the total number of humans and joint of a human body in a frame, respectively.

\textbf{Human features} are obtained from the human keypoints. we define it as $\textbf{P}_{t}^{n} \in \mathbb{R}^{D}$, and aggregate the information of $K$ joints of the $n^{\text{th}}$ person to get corresponding human features.

\textbf{Group features} are obtained from human features. It represents the features of a subgroup in the scene. We define this feature as $\textbf{G}_{t}^{m} \in \mathbb{R}^{D}$, where $M$ denotes the total number of subgroups. We aggregate human features from the $m^{\text{th}}$ subgroup to get this feature.

%After aggregating position information, velocity information, position embedding, coordinates embedding, and so on into those features, 
Then, we use a fully connected layer to unify multiple feature information into a fixed $D$ dimension:
\begin{equation}
f=w_{2}(\delta(w_{1}f+b_{1}))+b_{2}
\end{equation}
where $w_{1}$ and $w_{2}$ is learnable parameters, $b_1$ and $b_2$ are the bias vectors, and $\delta$ is the ReLU activation function. $f$ stands for one of the above four features.  

\subsection{Dynamic-composition Module}
% danni: 1. (和intro表达一致) 
The dynamic composition relationship between humans is discovered via spatial-temporal information encoding and mining. This module is mainly based on a tailored transformer, using the multi-head attention mechanism in both temporal and spatial dimensions, and strengthening the temporal and spatial features through embedding.

Before entering the encoder, the time and position are embedded in the input to enhance the spatio-temporal information. Spatial encoder and temporal encoder are used to extract and optimize features separately. 
\begin{equation}
\begin{aligned}
    \textbf{Dim}_S &:  \{ \textbf{P}_{1}, \textbf{P}_{2}, \dots, \textbf{P}_{N} \}, \textbf{P}_n = \{\textbf{P}_n^1, \textbf{P}_n^2, \dots, \textbf{P}_n^T\} 
    \\
    \textbf{Dim}_T &:  \{ \textbf{P}^1, \textbf{P}^2, \dots, \textbf{P}^T \}, \textbf{P}^t = \{\textbf{P}_1^t, \textbf{P}_2^t, \dots, \textbf{P}_N^t\}
\end{aligned}
\end{equation}
where the $\textbf{Dim}_S$ and $\textbf{Dim}_T$ represent spatial and temporal encoder respectively. $N$ and $T$ are the number of all persons and the number of frames we used in each clip. We design a circle structure to connect the spatial and temporal encoder to enhance the spatio-temporal features, taking the output of the temporal encoder as the input of the spatial encoder. We need to adjust the dimension of the output of the temporal encoder so that the spatial encoder can perform the calculation of the attention mechanism in the spatial dimension. By repeating the processing circle, our final message will be optimized in both the temporal and spatial dimensions. % danni: 从图上看一个 两个encoder形成了一个循环，先用一个总结把这个总体设计说一下～ 会清楚一点。 比如 We design a circle structure to connect sptical and temporal encoder to enhance the spatio-temporal features. 

The temporal encoder is designed to inform human features with temporal dynamical evolution clues and take $\textbf{P}_{n}^{t} \in \mathbb{R}^{D}$ which corresponds to the human features as input. %The $P_{t}^{n}$ corresponds to the human features in the above, but we need to transpose 
Then, the spatial dimension of input is transposed as the batch dimension. We utilize a multi-headed attention mechanism to explore the temporal relations and the formula is shown below:
\begin{equation}
\textbf{X}_{n} = \textbf{P}_{n} + \textbf{E}_\text{time}
\label{e_embed1}
\end{equation}
\begin{equation}
\label{e_before_att1}
\textbf{Q}_n=w_q \textbf{X}_{n}, \textbf{K}_n=w_k \textbf{X}_{n}, \textbf{V}_n=w_v \textbf{X}_{n}
\end{equation}
The $\textbf{X}_{n}$ here represents the $n$-th human features of $T$ frames, and $\textbf{E}_\text{time}$ denotes the embedding of pure timing information. $w_q$, $w_k$, $w_v$ are learnable parameters shaped $D\times D$. $\textbf{Q}_n$, $\textbf{K}_n$ and $\textbf{V}_n$ donate the query, key and value of the $n$-th human features of $T$ frames. Then, we use self-attention and softmax to update the values:
\begin{equation}
\label{e_attention}
\textbf{V}_n=\textrm{softmax}(\dfrac{\textbf{Q}_n {\textbf{K}_n}^\top}{\sqrt{D}})\textbf{V}_n+\textbf{V}_n
\end{equation}
The $\top$ represents the transpose of a matrix. Finally, a fully connected feed-forward network (FFN) in \cite{vaswani2017attention} is used:
\begin{equation}
\textbf{V}_n=\hat{w}_{2}\max (0, \hat{w}_{1} x + \hat{b}_{1})+\hat{b}_{2}
\end{equation}
where $\hat{w}_{1}$, $\hat{w}_{2}$, $\hat{b}_{1}$ and $\hat{b}_{2}$ are the parameters after the temporal encoder update. 

After that, we change the dimension of the $\textbf{V}_n$ from $N\times T\times D$ to $T\times N\times D$ as $V^t$ to meet the requirements of the spatial encoder.
Similarly,
% aims to explore spatial relations and refine features with spatial information,
the spatial encoder takes $\textbf{V}^t$ as input and views temporal dimension as the batch dimension, where $t$ denotes the $t$-th frame. We then refine it with spatial embedding information, which is represented by $E_\text{pos}$ in the following formula.
\begin{equation}
    \label{e_embed2}
        \textbf{V}^t=\textbf{V}^t+\textbf{E}_\text{pos}
\end{equation}
where $\textbf{V}^t$ represents the value of $t$-th frame including the feature of all $N$ individuals. $\textbf{E}_\text{pos}$ denotes the embedding of pure spatial information. 
\begin{equation}
\label{e_before_att2}
\textbf{Q}^t=\bar{w}^t \textbf{X}^t, \textbf{K}^t= \bar{w}^k X^t , \textbf{V}^t = \bar{w}_t X^t
\end{equation}
where $\textbf{Q}^t$, $\textbf{K}^t$ and $\textbf{V}^t$ are the query, key and value of the $t$-th frame features of all $N$ individuals. $\bar{w}_q$, $\bar{w}_k$, $\bar{w}_v$ are learnable parameters after the spatial encoder update.
\begin{equation}
\label{e_attention2}
\textbf{V}^t=\textrm{softmax}(\dfrac{\textbf{Q}^t {\textbf{K}^t}^\top}{\sqrt{D}})\textbf{V}^t+\textbf{V}^t
\end{equation}

Spatial encoder will process the temporal informed human features in the view of spatial position, which aims to explore the relation of all participants and the overall activity at a single frame. Since the temporal encoder uses human features of one person in $T$ frames and makes it more clarity and reliability, spatial encoder can utilize these refined features and get a more effective output representing human relations, making it not only aware of spatial neighbors but also the temporal dynamic features itself.

%Spatial and temporal information are processed simultaneously to get a complete information about the spatio-temporal relationship. The output of one encoder is used as the input of another and address them in other dimension, repeating the above process multiple times allows spatial and temporal information to fully collaborate and mix. This makes the temporal and spatial information no longer separate, but merge together into a more comprehensive and trustworthy feature. so the fusion of spatio-temporal information allows relation features to contain both location and action information, becoming a higher level of scene information.

A reorganization of spatially and temporally refined features is designed to reflect the representation of human relationships.
Ultimately, a matrix is used to represent the final relation information, where each element of the matrix is stitched together with the features of the two persons corresponding to its rows and columns.
\begin{equation}
\textbf{R}=
\left[
\begin{matrix}
 r_{1,1} & r_{1,2}& \cdots & r_{1,j}      \\
 r_{2,1}      & r_{2,2}      & \cdots & r_{2,j}      \\
 \vdots & \vdots & \ddots & \vdots \\
 r_{i,1}      & r_{i,2}      & \cdots & r_{i,j}      \\
\end{matrix}
\right] ,r_{i,j} = \left[
 V_{i}\, V_{j} 
\right]
\end{equation}

Before the stitching operation starts, we change the dimension of the $\textbf{V}_n$ from $T\times N\times D$ to $N\times T\times D$. The $\textbf{V}_{i}$ and $\textbf{V}_{j}$ mean the $i$-th and $j$-th person, $r$ denotes the final relation information. The matrix $\textbf{R}$ reflects the human relations between every two persons. 
% By using Spatial-Temporal Transformer, we explore more comprehensive relation features that take into account both spatial and temporal behavioral information of other people. 
Because coordinate information is included in all features and spatial location information is also injected several times during the optimization process, we have reason to believe that this matrix contains rich group distribution information, which forms the composition together with human relations.

\subsection{Dynamic-interaction Module}
\begin{figure}
    \centering
    \includegraphics[width=\columnwidth]{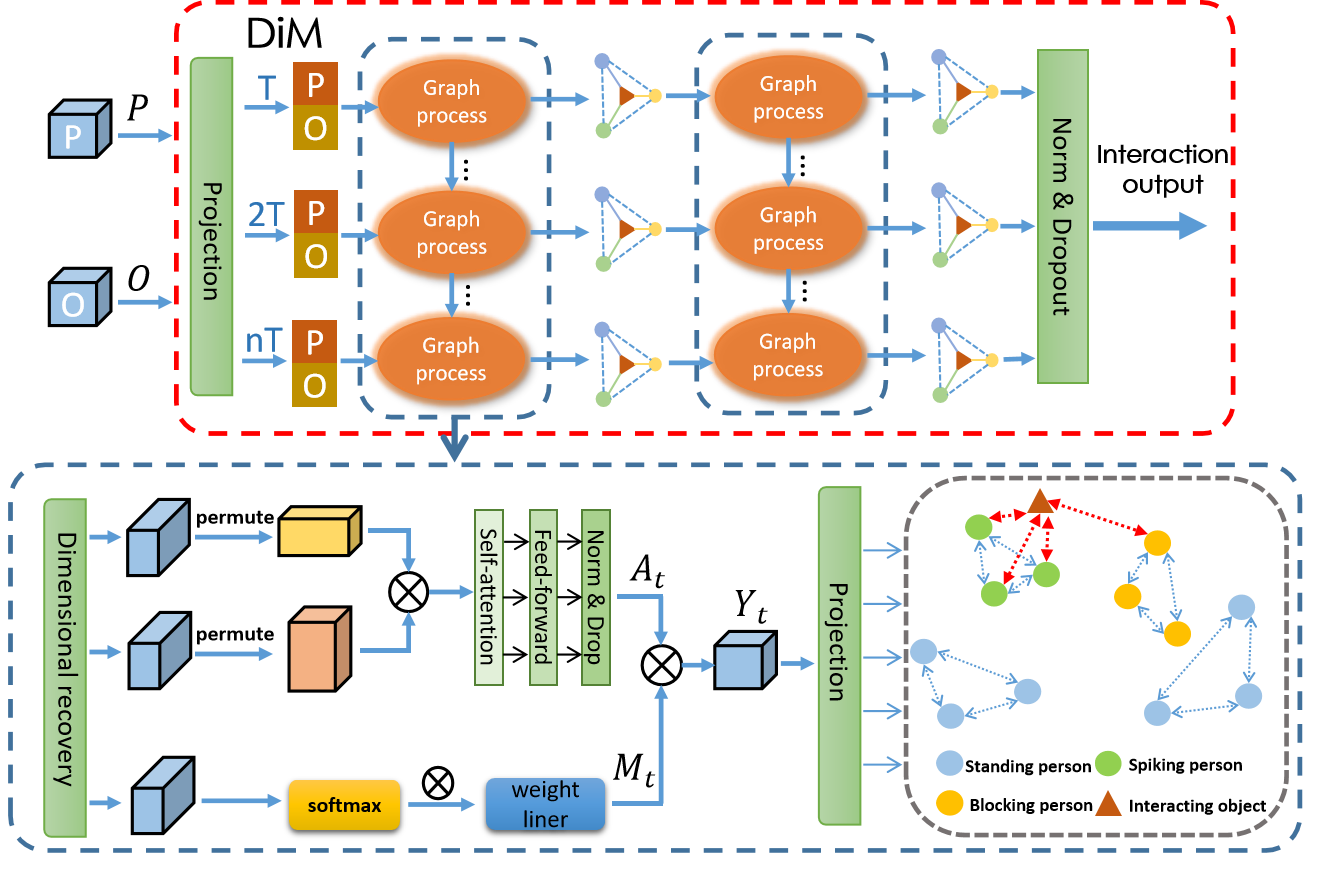}
    \caption{\label{fig:distribution-aware interaction}Our Dynamic-interaction Module (DiM) utilizes GCN and transformer to generate interaction information.} 
    \end{figure}
While extracting the relation information, another scene information is also worth our attention. 
This information is associated with both people and objects and is clearly important in group activity recognition, which we define as interaction. Since the number of people and objects in the real environment is often uncertain, and the interaction information is closer to a graph structure than a two-dimensional data structure, we use graph neural networks to model the interaction and the network structure is shown in Figure \ref{fig:distribution-aware interaction}. 

A novel graph that involves both human features and object features is designed to explore the interaction in an activity, using $\textbf{m}_{j}^{t}$ to denote a graph node with geometric features from either a human $\textbf{P}_{n}^{t}$ or an object $\textbf{O}_{n}^{t}$ at frame $t$. 

We utilize an adaptive adjacency matrix to represent the similarity of node features in our GCN, the similarity adjacency matrix $\textbf{A}_{t}$ is obtained from the dot product of the input and the swapped input, which represents if and how strong a connection exists between two nodes in the same frame $t$. The value of this matrix is calculated by the following formula:
\begin{equation}
\begin{aligned}
\textbf{A}&=\left \{ 
    \textbf{A}_{1}, \textbf{A}_{2}, \dots, \textbf{A}_{T} \}
    \right. \\
\textbf{a}_t(i,j)&= \text{Dropout} (m_i^\top) \text{Norm} (m_j)
\end{aligned}
\end{equation}
where $\textbf{Dropout}$ and $\textbf{Norm}$ are linear combinations of the dropout function and the normalization function. $\textbf{a}_t(i,j)$ is value of the $i$-th row and $j$-th column of matrix $\textbf{A}_{t}$ and $\textbf{A}$ means the similarity adjacency matrix of total $T$ frames.
The value $m_{i}$ means the features of a human or object at frame $t$.

After that, the $\textbf{SoftMax}$ activation function is applied on each row of $\textbf{A}_{t}$ to ensure the integration of all edge weights of a node equal to 1. To ensure the consistency of the matrix parameters at time $T$ and to take into account the timing information, we use a transformer to process all matrices at time $T$ and normalize them again with SoftMax. The final output can be obtained by the following:
\begin{equation}
\textbf{A}=\textrm{SoftMax}(\text{Encoder}(\textbf{A}))
\end{equation}
\begin{equation}
\textbf{Y}_{t} = \textbf{A}_{t} {\textbf{M}}_{t}
\end{equation}
where $\textbf{Y}_{t}$ denotes the output of our GCN of frame $t$. $\textbf{Encoder}$ is standard encoder layer based on \cite{vaswani2017attention}.

\subsection{Multi-level Dynamic Integration}
In the above approach, we get 6 different levels of information that describe the whole scene from local to global, from concrete to abstract, and from different perspectives. A Multi-Level Integration  Transformer is proposed to integrate and make full use of that information. The low-level information complements and improves the high-level information, and the different information is exchanged through the Transformer, and finally, better global information is obtained under a comprehensive consideration.

However, the representation of object features $\textbf{O}_{n}^{t}$, human keypoints $\textbf{S}_{n}^{t,k}$, human features $\textbf{P}_{n}^{t}$ and group features $\textbf{G}_{m}^{t}$ still lack temporal relationship. Therefore, we use a multi-layer perception to recombine and divide the features internally so that this information becomes a kind of feature representation within a time period. 
\begin{equation}
\begin{aligned}
   \overline{\textbf{O}}_{n} &= \text{proj}(\textbf{O}_{n}^{1}, \textbf{O}_{n}^{2},\dots,\textbf{O}_{n}^{T}),
    \\
    \overline{\textbf{S}}_{n}^k &= \text{proj}(\textbf{S}_{n}^{1,k}, \textbf{S}_{n}^{2,k},\dots,\textbf{S}_{n}^{T,k}),
    \\
    \overline{\textbf{P}}_{n} &= \text{proj}(\textbf{P}_{n}^{1},\textbf{P}_{n}^{2},\dots,\textbf{P}_{n}^{T}),
    \\
    \overline{\textbf{G}}_{m} &= \text{proj}(\textbf{G}_{m}^{1},\textbf{G}_{m}^{2},\dots,\textbf{G}_{m}^{T})
\end{aligned}
\end{equation}
We use four different layers of Transformer in the network for processing and optimization of multiple features. In order to carry out the transfer of information between different levels and to supplement the high-level information, we use Projection to transform the low-level information into higher-level information and fuse it with the high-level information to achieve the update of the high-level information. The projection here denotes a Feed Forward Network (FFN) used for converting one kind of information into another. The network structure diagram of our Multi-Level Integrate Transformer and the detailed description of each feature is shown in Figure \ref{fig:overview}.

\section{Experimantal Results and Analysis}
\subsection{Dataset}
\textbf{Volleyball dataset.} The Volleyball dataset \cite{ibrahim2018hierarchical} consists of multiple volleyball clips with a length of 41 frames. The middle frame of each clip contains bounding box coordinates, individual action labels, and group activity labels. Individual action labels contain 9 actions: setting, digging, falling, jumping, blocking, moving, spiking, waiting, and standing. Group activity labels contain 8 activities, namely right set, right pass, right spike, right winpoint, left set, left pass, left spike, and left winpoint. This dataset contains 55 volleyball videos with 4,830 labeled frames (3493/1337 for training/testing).  We employ the metrics of group activity accuracy and individual action accuracy, following previous work~\cite{li2021groupformer, gavrilyuk2020actor}

\textbf{Collective Activity Dataset.} The Collective Activity dataset \cite{choi2009they} consists of 44 videos. The middle frame of every 10 frames contains the bounding box coordinate annotations and individuals’ action labels. Group activity labels contain waiting, talking, queuing, crossing, and walking. We use 32 videos for training and 12 videos for testing, following previous works \cite{yuan2021spatio,qi2018stagnet}. We use group activity accuracy as our evaluation metrics.
%danni: 1. 这里的split容易描述吗，容易的话用一句话描述一下~ 便于别人复现你的方法～ 是不是需要加上 metrics description？

\begin{table}
\begin{center}
\small

\begin{tabular}{lcccccc}
\toprule
Model & keypoint & RGB & flow & VD &VD indiv. & CAD\\
\midrule
AT~\cite{gavrilyuk2020actor} &  & \checkmark & \checkmark & 93.0 & --& 92.8\\
GIRN~\cite{perez2022skeleton} & \checkmark & \checkmark & \checkmark & 94.0 & --& 95.2\\
Gavrilyuket al.~\cite{gavrilyuk2020actor} & \checkmark &  & \checkmark & 94.4 & \textbf{85.9} & 91.2 \\
GroupFormer~\cite{li2021groupformer} & \checkmark & \checkmark & \checkmark & 
\textbf{95.7} & 85.6 & \textbf{96.3} \\ \midrule
HDTM~\cite{ibrahim2016hierarchical} &  & \checkmark &  & 81.9 & -- & 91.5\\
CERN~\cite{shu2017cern} &  & \checkmark &  & 83.3 &  69.1 & 87.2\\
HRN~\cite{ibrahim2018hierarchical} &  & \checkmark &  & 89.5 & -- & -- \\
SSU~\cite{bagautdinov2017social} &  & \checkmark &  & 90.6 & 81.8 & --\\
stagNet~\cite{qi2018stagnet} &  & \checkmark &  & 89.3 & -- & 89.1\\
ARG~\cite{wu2019learning} &  & \checkmark &  & 92.5 &  83.0 & 91.0\\
HiGCIN~\cite{yan2020higcin} &  & \checkmark &  & 91.5 & -- & 93.4\\
DIN~\cite{yuan2021spatio} &  & \checkmark &  & 93.6 & -- & -- \\
CRM~\cite{azar2019convolutional} &  & \checkmark &  & 93.0 & -- & -- \\
\midrule
COMPOSER~\cite{zhou2022composer} & \checkmark &  &  & 93.7$\dag$ & -- & 94.1$\dag$\\
Ours & \checkmark &  &  & \textbf{95.3} & 85.4 & \textbf{94.4}\\
\bottomrule
\end{tabular}
\end{center}
\caption{Comparison with state of the art on the Volleyball dataset (VD) and Collective dataset in terms of Activity Acc.\%. Keypoint, RGB, and flow(optical flow) are three widely used information modalities and a check in the table means that they are used in the corresponding model. VD indiv represents the accuracy metrics of individual action predictions. $\dag$ indicates that the data here are from our replication results.}
\label{table:VDandCAD}
\end{table}

\subsection{Implementation details}
The human keypoints information we obtain can be described as $\textbf{F}_{t}^{n,k} \in \mathbb{R}^{\overline{D}}$. $\overline{D}=11 $, which contains absolute and relative coordinates, absolute and relative velocities, normalized coordinates, and keypoint types in 2 dimensions. %danni: 解释F_{t}^{n,j} R N J T D 的意思～（上次提的，如果在方法部分说了，这里就不管了）
The last dimension of all features we mentioned above has $D$ values, which equals 256. We select $T = 10$ frames in the clips for training and testing on both datasets, which contain 5 frames before the middle frames and 4 after the middle frames. The number of spatial and temporal encoder layers in our DcM is set as 3. The dimension of the FFN layer in all Transformer encoders is set as 1024, and the non-linear activation function is ReLU. The dropout rate of the Transformer encoder at each scale is set as 0.3.

We obtain human keypoints following \cite{gavrilyuk2020actor} and the person bounding box data is provided by \cite{sendo2019heatmapping}. The keypoints we use have 17 different types, and the person number of the Volleyball dataset is 12 while the Collective Activity is 13. When using the Volleyball dataset, the object's keypoints are from \cite{gavrilyuk2020actor}. To reduce the problem caused by noisy estimated keypoints, we use the temporal Object Keypoint Similarity (OKS) proposed in \cite{snower202015}.

In the training process, for both datasets, we adopt adam to learn the network parameters with a learning rate of 0.001. We set the weight decay to 0.001 and use a batch size of 384. The network is implemented by PyTorch on two NVIDIA Tesla P100 with a GPU of 12GB memory and trained for 60 epochs.

\subsection{Comparison with the State-of-the-Art}
In this subsection, We compare DynamicFormer with the state-of-the-art methods, since the two main datasets that are currently widely recognized in the group activity recognition field are the collective dataset and the volleyball dataset.

\textbf{Volleyball dataset.} The results of our comparison with other SOTA methods are listed in Table \ref{table:VDandCAD}. Our method surpasses many previous methods, especially those which use RGB-only and keypoints-only information. And we achieve the performance of state-of-the-art though most of them use additional information like optical flow. % danni: 1. (give some examples) like what information?
For the RGB-only methods, compared to the optimal methods CRM and DIN, DynamicFormer makes full use of group composition and object-human interaction information, allowing for accuracy improvements of 2.3\% and 1.7\% respectively. 

For the multi-modal approach, it is worth noting that our approach outperforms AT, GIRN, and COMPOSER by approximately 2.3\%, 1.3\%, and 0.7\% in terms of accuracy of group activity identification, despite the increased information and computational complexity of these approaches. % danni: 2. 这里this approach指的是我们的方法还是AT GIRN COMPOSER -- these approaches?
It's because our model captures and exploits the relation information among humans and interaction information between humans and objects. Although there is a slight difference (0.4\%) in our approach compared to GroupFormer, 
% and a clustered attention mechanism to dynamically divide individuals into multiple clusters, 
our method considers interaction to discern the importance of humans simply and a more comprehensive group composition information to understand the scene. The main reason for the difference is that GroupFormer utilizes additional optical flow information and divide individuals into multiple clusters, which further improved performance. We do not use optical flow information because optical flow contains rich motion information, but does not help us to investigate the interactions.
% danni: 3. 这里的分析还不够说明两个方法的优劣之处，需要分析推测为什么groupformer的效果好一些。
% (我们方法比他好的地方). 
\begin{figure}
    \centering
    \includegraphics[width=\columnwidth]{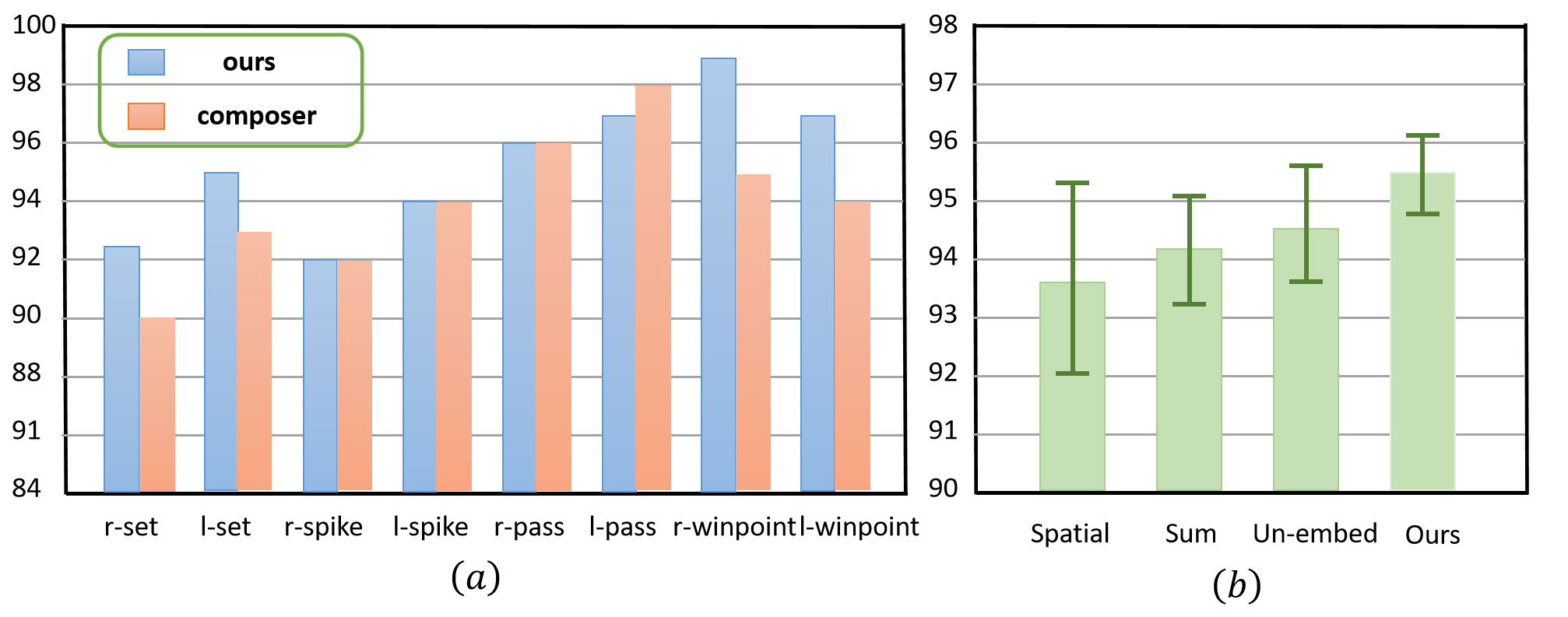}
    \caption{\label{fig:matrices} (a) shows the comparison of the recognition accuracy of composer~\cite{zhou2022composer} and ours on different classes. (b) shows the group recognition accuracy and stability achieved using different composition modules.}
\end{figure}

We also provide the accuracy of each class in our results in Figure \ref{fig:matrices} (a), our model performs well in all categories. For r-winpoint and l-winpoint, the member of the defensive side does not successfully stop the ball, making the ball-human interaction during this period relatively weak and easier to discern.
%(分析一下这个类为什么识别效果好)，
 For r-set and r-spike, most failures can be a result of highly similar actions of key persons and human composition in some clips,
%(识别不好的原因)
but our collaborating interaction and composition information helps to identify some subtle differences in human composition and key actions. Also, our method outperforms or equals composer in almost all categories, which further demonstrates the superiority of our method.
%(我们的***帮助解决了部分问题，可以让这两个类达到了一个可观的识别准确度)
% danni: 4. 这里能不能加上groupformer方法对不同类的识别准确率？进行对比分析，找到我们方法和groupformer的优劣之处。 (b)这个图好像没有分析

\textbf{Collective dataset.} This dataset has no object and we can not use the human-object interaction information in this dataset evaluation. We change the interaction GCN block to generate supplementary information to our composition information.
% We change the interaction GCN block and use a linear-like structure to reorganize the multi-level transformer as supplementary information in an alternative to relation information.
% Then this block uses the joints, individuals, relations, and groups as progressive levels of information to adapt this dataset.   
The results of our model compared with other SOTA methods are listed in Table \ref{table:VDandCAD}, which shows that in addition to the interaction block, our composition block and multi-level information aggregation methods also perform well for group activity recognition. 

Although the analysis of this dataset does not use human-object interaction information, our method still achieves the best results compared to RGB-Only. For the multi-modal method AT, without human-object interaction information, our method still achieves 1.6\% improvement since we consider the comprehensive composition, including human relations and group distribution in a global view. %(补充我们方法的优点).
Compared to GIRN, we can achieve a significant improvement on the Volleyball dataset but have a slight accuracy difference without human-object interaction information, which validates that discovering human-object interaction information can significantly improve group activity recognition. 
\begin{table*}
\centering
    \small
    \begin{tabular}{lcc}
    \multicolumn{2}{c}{(a)}\\
    \toprule
    Manner & Group Activity  \\
    \midrule
    Baseline & 93.1  \\
    Baseline \textit{w/} Spatial & 93.6 \\
    Baseline \textit{w/} Sum & 94.2 \\
    Baseline \textit{w/} Un-embed & 94.4 \\
    \midrule
    Ours & 95.3 \\
    \bottomrule
    \end{tabular}
     \quad  \quad
    \begin{tabular}{cc}
    \multicolumn{2}{c}{(b)}\\
    \toprule
    Manner & Group Activity  \\
    \midrule
    None-ball & 92.7  \\
     None-trans & 93.6 \\
    Erase & 91.9 \\
    Ours & 95.3 \\
    \bottomrule
    \end{tabular}
    \quad  \quad
    \begin{tabular}{cc}
    \multicolumn{2}{c}{(c)}\\
    \toprule
    Manner &Group Activity  \\
    \midrule
    Linear &94.3  \\
    Parallel &94.0 \\
    Ours &95.3 \\
    \bottomrule
    \end{tabular}
    \caption{(a), (b) represent different composition and interaction features extractors respectively. (c) shows the results of different integrated orders.}
    \label{table:ablation}
\end{table*}
\subsection{Ablation Studies}
In this subsection, we perform detailed ablation studies on the Volleyball dataset to explore the contribution and role of each part of our model using group activity recognition accuracy as the evaluation metric.

% \begin{table}
% \begin{center}
% \small
% \setlength{\tabcolsep}{6.4mm}
% \begin{tabular}{lcc}
% \toprule
% Manner & Group Activity  \\
% \midrule
% Baseline & 93.1  \\
% Baseline \textit{w/} Spatial & 93.6 \\
% Baseline \textit{w/} Sum & 94.2 \\
% Baseline \textit{w/} Un-embed & 94.4 \\
% \midrule
% Ours & 95.3 \\
% \bottomrule
% \end{tabular}
% \end{center}
% \caption{Exploration of different relation feature extractor}
% \label{table:relation}
% \end{table}

\textbf{Variations of Composition.}
We begin our experiments by studying the effects of the different ways to combine spatial-temporal information, which is also able to verify the advance of our dynamic composition module design. As shown in Table~\ref{table:ablation} (a), we use four variants of our DcM for replacement and compare their results. The settings are (1) Baseline: We replace spatial and temporal encoder with an FC layer after the time and position embedding. (2) Spatial manner: This variant uses only a spatial composition with embedding. (3) Sum manner: This variant involves both spatial and temporal encoders but simply fuses them with direct sum operation after two decoder layers, instead of using a circle design. (4) Un-embed manner: This variant uses the same spatial-temporal transformer without time and position embedding. 
% danni: 7. 没理解这几个模块和框架图里如何对应？ 框架图-方法描述-这里的描述 保持一致

As shown in Table~\ref{table:ablation} (a), using spatial information (Spatial manner) achieves 0.5\% improvement in Baseline, which shows that static human features and group spatial distribution have some but not a significant contribution to the GAR task. Simply fusing temporal information considering dynamic composition along time (Sum manner) improves 0.6\% than spatial-only. It indicates that encoding both spatial and temporal information introduces 
% the dynamics of human actions and spatial distribution, and therefore makes our model aware of 
dynamic composition information and performs 1.1\% better than baseline. The Un-embed improves by 0.2\% over a simple fusion of temporal information (Sum manner), which shows that the iterative information fusion of spatial and temporal relationship might be helpful for the model to perceive dynamic group composition.
% perceive dynamic group distribution and changing group actions and the connection between them.
Our DcM gained 0.9\% improvement over Un-embed, indicating that embedding time and positions to original human features improve the model's perception of dynamic group spatial relationships and human actions.

To sum up, our DcM can improve the performance of the spatial-only method from 93.6\% to 95.3\%, achieving a 1.1\% improvement over simply summing the results of the spatial-temporal transformer and 0.9\% improvement over Un-embed. It shows the valuable role of each part of the composition module. As shown in Figure~\ref{fig:matrices}(b), our DcM exhibits not only higher recognition accuracy, but also stronger stability in the recognition of different classes, which further illustrates that our approach can be useful for capturing dynamic composition features.
% Totally, our DcM %我们最终使用的方式}  
%  models dynamic composition features, which includes relation and person locations and can collaborate well with interactions.

\textbf{Variations of Interaction.}
To verify the effectiveness of the DiM, we present some different settings to replace our GCN network. (1) None-ball manner: This manner slightly alters our Interaction GCN's network structure to ignore the enlightening ball information. (2) None-trans manner: It omits the transformer used internally in our Interaction GCN. (3) Erase manner: The erase manner erases the GCN used in our DiM. 

All the settings of those variants are the same and the results are listed in Table \ref{table:ablation} (b). 
The 3.4\% increase achieved by our full method compared to the Erase method is strong evidence of the critical role of human-object interaction information in group activity recognition.
For None-trans, the loss of 1.7\% compared to the full method proves that our transformer inside the GCN can actually inject temporal information into our features and improve its representation. 
Compared to the full method, the accuracy of the None-ball manner is reduced by 2.6\%, because the None-ball manner can be seen as degenerating into the modeling of human relation information, which complements our relation transformer in a sense but ignore the more critical interaction information.
% \begin{table}
% \begin{center}
% \small
% \setlength{\tabcolsep}{5mm}
% \begin{tabular}{clcc}
% \toprule
% Modal & Manner & Group Activity  \\
% \midrule
% &None-ball & 92.7  \\
% \multirow{2}{*}{keypoint} &None-trans & 93.6 \\
% &Erase & 91.9 \\
% \cmidrule{2-3}
% &Ours & 95.3 \\
% \bottomrule
% \end{tabular}
% \end{center}
% \caption{Exploration of different interaction feature extractor}
% \label{table:interaction}
% \end{table}

\begin{figure}
    \centering
    \includegraphics[width=0.95\columnwidth]{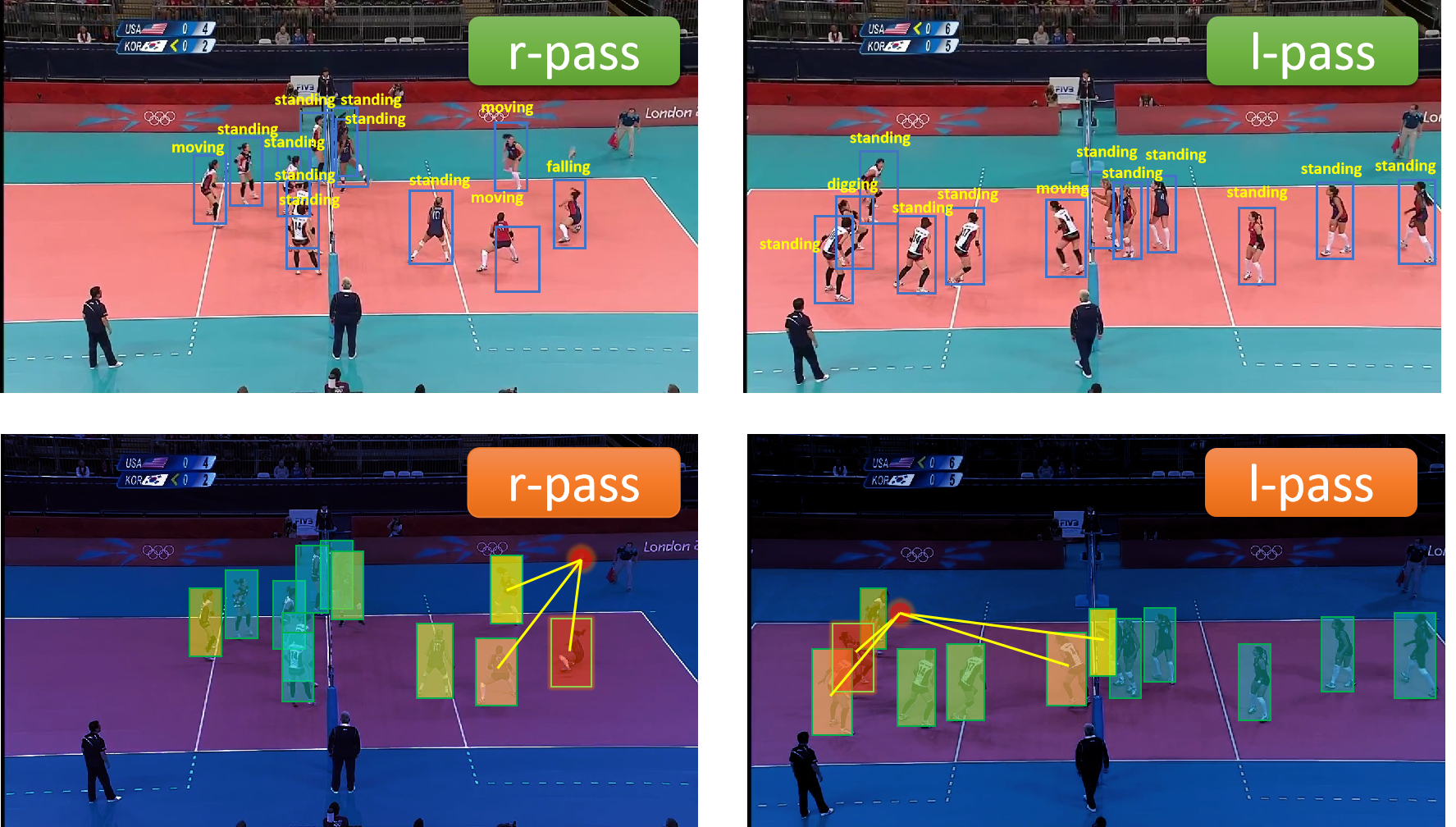}
    \caption{\label{fig:inter_visualization} Visualization of the effect of our DiM. The first row contains the bounding box and the groundtruth labels of person actions and group activity.}
\end{figure}

\textbf{Integration Efficiency.} Since our Dynamic Ordered Integration takes different levels of features as input, we present several settings to analyze impacts on the final result in Table~\ref{table:ablation} (c). (1) Linear manner: This method considers interaction, relation, and group as different levels of information, and uses the transformer in this order to aggregate and fuse the information layer by layer. (2) Parallel manner: This method considers the three types of features as the same level of information and fuse them directly. 
%Our model uses a bottom-up structure, treating relations and interactions as the same level of information, and eventually merging them with higher-level group information. We explored this structure using two variants.

As shown in Table \ref{table:ablation} (c), Linear and Parallel structures have achieved significant results. However, those structures ignore multi-level information interaction. Combining these two structures allows for an efficient ordering combination of multi-level information. We suspect that the reason for this may be twofold: (1) The main role of our multi-level transformer structure is to transform and summarize the various information previously obtained, so despite the change in structure, its main feature information is not lost. (2) The aggregation structure corresponding to the hierarchy of information itself can maximize the meaning and merit of information and produce a better representation of the whole scene, which will improve our final group activity accuracy.
% \begin{table}
% \begin{center}
% \small
% \setlength{\tabcolsep}{6mm}
% \begin{tabular}{clcc}
% \toprule
% Modal &Manner &Group Activity  \\
% \midrule
% &Linear &94.3  \\
% \multirow{1}{*}{keypoint} &Parallel &94.0 \\
% \cmidrule{2-3}
% & Ours &95.3 \\
% \bottomrule
% \end{tabular}
% \end{center}
% \caption{Exploration of different interaction feature extractor}
% \label{table:mlti-level}
% \end{table}

\begin{figure}
    \centering
    \includegraphics[width=0.95\columnwidth]{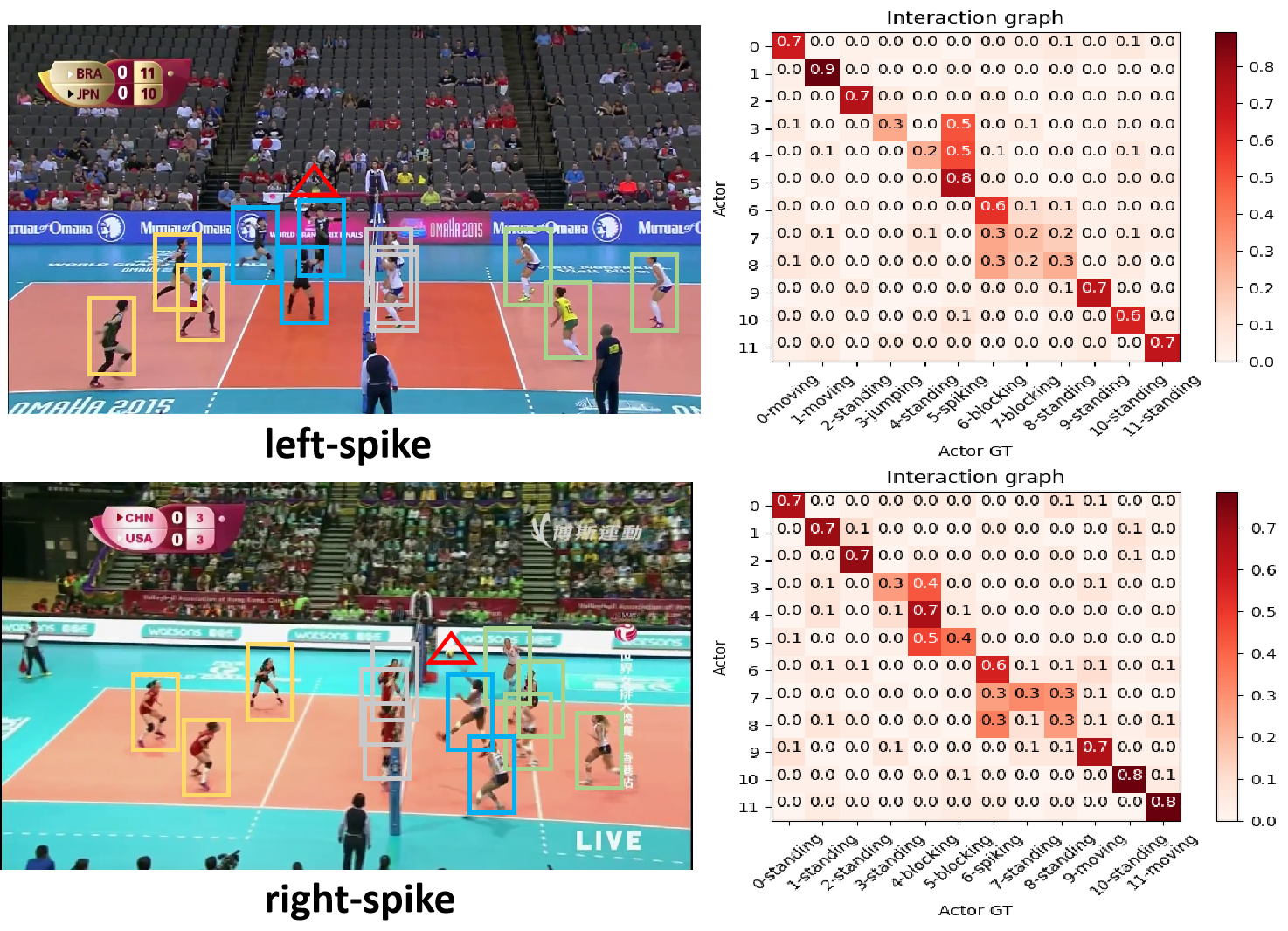}
    \caption{\label{fig:inter_graph} Visualization of learned composition and interactions. Boxes of the same color indicate a strong relationship between those people. The matrix on the right shows the classification of individuals with ground-truth individual labels. The person who has the max column sum of the matrix is the most important person and was marked with a red triangle.
    }
    \end{figure}
% The rows and columns in the interaction graph represent the people in the scene and the ground truth of the corresponding action of the people, respectively. Human-object interactions can have an impact on the behavioral representations of others, causing them to deviate to some extent. Therefore, the column with the largest sum of numbers in the figure represents the most influential localities in the whole scene, which just corresponds to the person who is interacting with an object, spiking and blocking a person in the graph.
\subsection{Visualization}
Figure~\ref{fig:inter_visualization} illustrates the importance of introducing the interaction. The second row reflects our DiM ability to identify the more important one. The different color shows different degree of importance. Also, the yellow lines represent who interacts more closely with the key object.

Figure~\ref{fig:inter_graph} represents the role of our composition and interaction. Since composition constructs human relations, and interaction emphasizes the important part, the person who has a close relation to the key person may be assimilated, making this information more visible in the global scene, which further promotes group activity recognition.

\section{Conclusion}
%还没修改
This paper has presented an efficient and comprehensive approach for inferring group activity in a complex multi-person scene, which is mainly based on the transformer and takes the keypoint-only modality as input. We first use a novel transformer called DiM to exploit the group composition in spatial and temporal dimensions simultaneously. Moreover, we utilize a DcM to model human-object interaction and explore its special and indispensable role in activity reasoning. 
 We evaluate our proposed model on two benchmarks and establish new state-of-the-art results.

% keywords can be removed
%\keywords{First keyword \and Second keyword \and More}

\bibliographystyle{unsrt}

\begin{thebibliography}{10}

\bibitem{deng2016structure}
Zhiwei Deng, Arash Vahdat, Hexiang Hu, and Greg Mori.
\newblock Structure inference machines: Recurrent neural networks for analyzing
  relations in group activity recognition.
\newblock In {\em Proceedings of the IEEE conference on computer vision and
  pattern recognition}, pages 4772--4781, 2016.

\bibitem{wang2018appearance}
Limin Wang, Wei Li, Wen Li, and Luc Van~Gool.
\newblock Appearance-and-relation networks for video classification.
\newblock In {\em Proceedings of the IEEE conference on computer vision and
  pattern recognition}, pages 1430--1439, 2018.

\bibitem{tran2015learning}
Du~Tran, Lubomir Bourdev, Rob Fergus, Lorenzo Torresani, and Manohar Paluri.
\newblock Learning spatiotemporal features with 3d convolutional networks.
\newblock In {\em Proceedings of the IEEE international conference on computer
  vision}, pages 4489--4497, 2015.

\bibitem{simonyan2014two}
Karen Simonyan and Andrew Zisserman.
\newblock Two-stream convolutional networks for action recognition in videos.
\newblock {\em Advances in neural information processing systems}, 27, 2014.

\bibitem{li2021groupformer}
Shuaicheng Li, Qianggang Cao, Lingbo Liu, Kunlin Yang, Shinan Liu, Jun Hou, and
  Shuai Yi.
\newblock Groupformer: Group activity recognition with clustered
  spatial-temporal transformer.
\newblock In {\em Proceedings of the IEEE/CVF International Conference on
  Computer Vision}, pages 13668--13677, 2021.

\bibitem{zhou2022composer}
Honglu Zhou, Asim Kadav, Aviv Shamsian, Shijie Geng, Farley Lai, Long Zhao,
  Ting Liu, Mubbasir Kapadia, and Hans~Peter Graf.
\newblock Composer: Compositional reasoning of group activity in videos with
  keypoint-only modality.
\newblock {\em Proceedings of the 17th European Conference on Computer Vision
  (ECCV 2022)}, 2022.

\bibitem{azar2019convolutional}
Sina~Mokhtarzadeh Azar, Mina~Ghadimi Atigh, Ahmad Nickabadi, and Alexandre
  Alahi.
\newblock Convolutional relational machine for group activity recognition.
\newblock In {\em Proceedings of the IEEE/CVF Conference on Computer Vision and
  Pattern Recognition}, pages 7892--7901, 2019.

\bibitem{ibrahim2018hierarchical}
Mostafa~S Ibrahim and Greg Mori.
\newblock Hierarchical relational networks for group activity recognition and
  retrieval.
\newblock In {\em Proceedings of the European conference on computer vision
  (ECCV)}, pages 721--736, 2018.

\bibitem{bagautdinov2017social}
Timur Bagautdinov, Alexandre Alahi, Fran{\c{c}}ois Fleuret, Pascal Fua, and
  Silvio Savarese.
\newblock Social scene understanding: End-to-end multi-person action
  localization and collective activity recognition.
\newblock In {\em Proceedings of the IEEE conference on computer vision and
  pattern recognition}, pages 4315--4324, 2017.

\bibitem{li2017sbgar}
Xin Li and Mooi Choo~Chuah.
\newblock Sbgar: Semantics based group activity recognition.
\newblock In {\em Proceedings of the IEEE international conference on computer
  vision}, pages 2876--2885, 2017.

\bibitem{shu2017cern}
Tianmin Shu, Sinisa Todorovic, and Song-Chun Zhu.
\newblock Cern: confidence-energy recurrent network for group activity
  recognition.
\newblock In {\em Proceedings of the IEEE conference on computer vision and
  pattern recognition}, pages 5523--5531, 2017.

\bibitem{wu2019learning}
Jianchao Wu, Limin Wang, Li~Wang, Jie Guo, and Gangshan Wu.
\newblock Learning actor relation graphs for group activity recognition.
\newblock In {\em Proceedings of the IEEE/CVF Conference on computer vision and
  pattern recognition}, pages 9964--9974, 2019.

\bibitem{ehsanpour2020joint}
Mahsa Ehsanpour, Alireza Abedin, Fatemeh Saleh, Javen Shi, Ian Reid, and Hamid
  Rezatofighi.
\newblock Joint learning of social groups, individuals action and sub-group
  activities in videos.
\newblock In {\em Computer Vision--ECCV 2020: 16th European Conference,
  Glasgow, UK, August 23--28, 2020, Proceedings, Part IX 16}, pages 177--195.
  Springer, 2020.

\bibitem{nakatani2021group}
Chihiro Nakatani, Kohei Sendo, and Norimichi Ukita.
\newblock Group activity recognition using joint learning of individual action
  recognition and people grouping.
\newblock In {\em 2021 17th International Conference on Machine Vision and
  Applications (MVA)}, pages 1--5. IEEE, 2021.

\bibitem{YanSYTT22}
Rui Yan, Xiangbo Shu, Chengcheng Yuan, Qi~Tian, and Jinhui Tang.
\newblock Position-aware participation-contributed temporal dynamic model for
  group activity recognition.
\newblock {\em {IEEE} Trans. Neural Networks Learn. Syst.}, 33(12):7574--7588,
  2022.

\bibitem{DengVHM16}
Zhiwei Deng, Arash Vahdat, Hexiang Hu, and Greg Mori.
\newblock Structure inference machines: Recurrent neural networks for analyzing
  relations in group activity recognition.
\newblock In {\em 2016 {IEEE} Conference on Computer Vision and Pattern
  Recognition, {CVPR} 2016, Las Vegas, NV, USA, June 27-30, 2016}, pages
  4772--4781, 2016.

\bibitem{ji20123d}
Shuiwang Ji, Wei Xu, Ming Yang, and Kai Yu.
\newblock 3d convolutional neural networks for human action recognition.
\newblock {\em IEEE transactions on pattern analysis and machine intelligence},
  35(1):221--231, 2012.

\bibitem{ibrahim2016hierarchical}
Mostafa~S Ibrahim, Srikanth Muralidharan, Zhiwei Deng, Arash Vahdat, and Greg
  Mori.
\newblock A hierarchical deep temporal model for group activity recognition.
\newblock In {\em Proceedings of the IEEE conference on computer vision and
  pattern recognition}, pages 1971--1980, 2016.

\bibitem{kipf2016semi}
Thomas~N Kipf and Max Welling.
\newblock Semi-supervised classification with graph convolutional networks.
\newblock {\em arXiv preprint arXiv:1609.02907}, 2016.

\bibitem{kipf2018neural}
Thomas Kipf, Ethan Fetaya, Kuan-Chieh Wang, Max Welling, and Richard Zemel.
\newblock Neural relational inference for interacting systems.
\newblock In {\em International conference on machine learning}, pages
  2688--2697. PMLR, 2018.

\bibitem{wu2020comprehensive}
Zonghan Wu, Shirui Pan, Fengwen Chen, Guodong Long, Chengqi Zhang, and S~Yu
  Philip.
\newblock A comprehensive survey on graph neural networks.
\newblock {\em IEEE transactions on neural networks and learning systems},
  32(1):4--24, 2020.

\bibitem{xu2018powerful}
Keyulu Xu, Weihua Hu, Jure Leskovec, and Stefanie Jegelka.
\newblock How powerful are graph neural networks?
\newblock {\em arXiv preprint arXiv:1810.00826}, 2018.

\bibitem{hu2020progressive}
Guyue Hu, Bo~Cui, Yuan He, and Shan Yu.
\newblock Progressive relation learning for group activity recognition.
\newblock In {\em Proceedings of the IEEE/CVF Conference on Computer Vision and
  Pattern Recognition}, pages 980--989, 2020.

\bibitem{vaswani2017attention}
Ashish Vaswani, Noam Shazeer, Niki Parmar, Jakob Uszkoreit, Llion Jones,
  Aidan~N Gomez, {\L}ukasz Kaiser, and Illia Polosukhin.
\newblock Attention is all you need.
\newblock {\em Advances in neural information processing systems}, 30, 2017.

\bibitem{wang2018videos}
Xiaolong Wang and Abhinav Gupta.
\newblock Videos as space-time region graphs.
\newblock In {\em Proceedings of the European conference on computer vision
  (ECCV)}, pages 399--417, 2018.

\bibitem{li2018videolstm}
Zhenyang Li, Kirill Gavrilyuk, Efstratios Gavves, Mihir Jain, and Cees~GM
  Snoek.
\newblock Videolstm convolves, attends and flows for action recognition.
\newblock {\em Computer Vision and Image Understanding}, 166:41--50, 2018.

\bibitem{gavrilyuk2020actor}
Kirill Gavrilyuk, Ryan Sanford, Mehrsan Javan, and Cees~GM Snoek.
\newblock Actor-transformers for group activity recognition.
\newblock In {\em Proceedings of the IEEE/CVF Conference on Computer Vision and
  Pattern Recognition}, pages 839--848, 2020.

\bibitem{perez2022skeleton}
Mauricio Perez, Jun Liu, and Alex~C Kot.
\newblock Skeleton-based relational reasoning for group activity analysis.
\newblock {\em Pattern Recognition}, 122:108360, 2022.

\bibitem{choi2009they}
Wongun Choi, Khuram Shahid, and Silvio Savarese.
\newblock What are they doing?: Collective activity classification using
  spatio-temporal relationship among people.
\newblock In {\em 2009 IEEE 12th international conference on computer vision
  workshops, ICCV Workshops}, pages 1282--1289. IEEE, 2009.

\bibitem{yuan2021spatio}
Hangjie Yuan, Dong Ni, and Mang Wang.
\newblock Spatio-temporal dynamic inference network for group activity
  recognition.
\newblock In {\em Proceedings of the IEEE/CVF International Conference on
  Computer Vision}, pages 7476--7485, 2021.

\bibitem{qi2018stagnet}
Mengshi Qi, Jie Qin, Annan Li, Yunhong Wang, Jiebo Luo, and Luc Van~Gool.
\newblock stagnet: An attentive semantic rnn for group activity recognition.
\newblock In {\em Proceedings of the European Conference on Computer Vision
  (ECCV)}, pages 101--117, 2018.

\bibitem{yan2020higcin}
Rui Yan, Lingxi Xie, Jinhui Tang, Xiangbo Shu, and Qi~Tian.
\newblock Higcin: Hierarchical graph-based cross inference network for group
  activity recognition.
\newblock {\em IEEE transactions on pattern analysis and machine intelligence},
  2020.

\bibitem{sendo2019heatmapping}
Kohei Sendo and Norimichi Ukita.
\newblock Heatmapping of people involved in group activities.
\newblock In {\em 2019 16th International Conference on Machine Vision
  Applications (MVA)}, pages 1--6. IEEE, 2019.

\bibitem{snower202015}
Michael Snower, Asim Kadav, Farley Lai, and Hans~Peter Graf.
\newblock 15 keypoints is all you need.
\newblock In {\em Proceedings of the IEEE/CVF Conference on Computer Vision and
  Pattern Recognition}, pages 6738--6748, 2020.

\end{thebibliography}

\end{document}